\documentclass[twocolumn]{article}

\usepackage{times}
\usepackage{graphicx} 
\usepackage{epstopdf}
\usepackage{subfigure} 
\usepackage{natbib}
\usepackage[boxed]{algorithm}
\usepackage{algorithmic}
\usepackage{hyperref}
\usepackage{framed}

\usepackage{amsmath}
\usepackage{array}
\usepackage{stackengine}

\usepackage[letterpaper, margin=1in, top=0.5in]{geometry}

\begin{document}

\newcommand{\jac}[2]{\frac{\partial #1}{\partial #2}}
\newcommand{\xhat}{\widehat{x}}
\newcommand{\yhat}{\widehat{y}}
\newcommand{\zhat}{\widehat{z}}
\newcommand{\vxhat}{\widehat\mathrm{x}}
\newcommand{\vzhat}{\widehat\mathrm{z}}
\newcommand{\setX}{\mathcal{X}}
\newcommand{\setB}{\mathcal{B}}
\newcommand{\E}{\text{E}}
\newcommand{\Var}{\text{Var}}
\newcommand{\Cov}{\text{Cov}}
\newcommand{\Fhat}{\widehat{F}}
\newcommand{\Thetahat}{\widehat{\Theta}}
\newcommand{\Norm}{\text{Norm}}
\newcommand{\BatchNorm}{\text{BN}}
\newcommand{\kk}{{(k)}}
\newcommand{\vx}{\mathrm{x}}
\newcommand{\vy}{\mathrm{y}}
\newcommand{\vz}{\mathrm{z}}
\newcommand{\vb}{\mathrm{b}}
\newcommand{\vu}{\mathrm{u}}
\newcommand{\comt}{// }
\renewcommand{\algorithmiccomment}[1]{\comt #1}
\newcommand{\BN}[2]{\text{BN}_{#2}(#1)}
\renewcommand{\algorithmicrequire}{\textbf{Input:}}
\renewcommand{\algorithmicensure}{\textbf{Output:}}
\newcommand{\mils}{\cdot 10^6}
\newcommand{\netw}[1]{{\sl #1}}
\newcommand{\norig}{\text{\sl N}}
\newcommand{\ntrain}{\norig_\mathrm{BN}^\mathrm{tr}}
\newcommand{\ninf}{\norig_\mathrm{BN}^\mathrm{inf}}
\setcitestyle{authoryear,round,citesep={;},aysep={,},yysep={;}}
\renewcommand{\cite}[1]{\citep{#1}}

\title{Batch Normalization: Accelerating Deep Network Training by Reducing Internal Covariate Shift}

\author{Sergey Ioffe \\Google Inc., {\sl sioffe@google.com} \and
Christian Szegedy \\Google Inc., {\sl szegedy@google.com}}

\date{}

\maketitle

\begin{abstract}

Training Deep Neural Networks is complicated by the fact that the
distribution of each layer's inputs changes during training, as the
parameters of the previous layers change.  This slows down the
training by requiring lower learning rates and careful parameter
initialization, and makes it notoriously hard to train models with
saturating nonlinearities.  We refer to this phenomenon as {\em
  internal covariate shift}, and address the problem by normalizing
layer inputs.  Our method draws its strength from making normalization
a part of the model architecture and performing the normalization {\em
  for each training mini-batch}.  Batch Normalization allows us to use
much higher learning rates and be less careful about
initialization. It also acts as a regularizer, in some cases
eliminating the need for Dropout.  Applied to a state-of-the-art image
classification model, Batch Normalization achieves the same accuracy
with 14 times fewer training steps, and beats the original model by a
significant margin. 
Using an ensemble of batch-normalized networks, we
improve upon the best published result on ImageNet classification:
reaching 4.9\% top-5 validation error (and 4.8\% test error), exceeding the accuracy
of human raters.

\end{abstract}

\section{Introduction}

Deep learning has dramatically advanced the state of the art in vision, speech,
and many other areas. Stochastic gradient descent (SGD) has proved to be an effective way of training
deep networks, and SGD variants such as momentum \cite{momentum} and Adagrad \cite{adagrad} have been used to
achieve state of the art performance. SGD optimizes the parameters $\Theta$ of
the network, so as to minimize the loss $$\Theta = \arg \min_\Theta
\frac{1}{N}\sum_{i=1}^N \ell(\vx_i, \Theta)$$ where
$\vx_{1\ldots N}$ is the training data set.  With SGD, the training
proceeds in steps, and at each step we consider a {\em mini-batch}
$\vx_{1\ldots m}$ of size $m$. The mini-batch is used
to approximate the gradient of the loss function with respect to the parameters,
by computing $$\frac{1}{m} \jac{ \ell(\vx_i, \Theta)}{ \Theta}.$$ Using
mini-batches of examples, as opposed to one example at a time, is helpful in
several ways. First, the gradient of the loss over a mini-batch is an estimate
of the gradient over the training set, whose quality  improves
as the batch size increases. Second, computation over a batch can be much more
efficient than $m$ computations for individual examples, due to the parallelism
afforded by the modern computing platforms.

While stochastic gradient is simple and effective, it requires careful tuning of
the model hyper-parameters, specifically the learning rate used in optimization,
as well as the initial values for the model parameters. The training is
complicated by the fact that the inputs to each layer are affected by the
parameters of all preceding layers -- so that
small changes to the network parameters amplify as the network becomes
deeper.

The change in the distributions of layers' inputs presents a problem
because the layers need to continuously adapt to the new
distribution. When the input distribution to a learning system
changes, it is said to experience {\em covariate shift}
\cite{covariate-shift}. This is typically handled via domain
adaptation \cite{domain-adaptation-survey}. However, the notion of
covariate shift can be extended beyond the learning system as a whole,
to apply to its parts, such as a sub-network or a layer. Consider a
network computing $$\ell = F_2(F_1(\vu, \Theta_1), \Theta_2)$$ where
$F_1$ and $F_2$ are arbitrary transformations, and the parameters
$\Theta_1, \Theta_2$ are to be learned so as to minimize the loss
$\ell$.  Learning $\Theta_2$ can be viewed as if the inputs
$\vx=F_1(\vu,\Theta_1)$ are fed into the sub-network
$$\ell = F_2(\vx, \Theta_2).$$ For example, a gradient descent step
$$\Theta_2\leftarrow \Theta_2 - \frac{\alpha}{m}\sum_{i=1}^m
\jac{F_2(\vx_i,\Theta_2)}{\Theta_2}$$ (for batch size $m$ and learning
rate $\alpha$) is exactly equivalent to that for a stand-alone network
$F_2$ with input $\vx$.  Therefore, the input distribution properties
that make training more efficient -- such as having the same
distribution between the training and test data -- apply to training
the sub-network as well.  As such it is advantageous for the
distribution of $\vx$ to remain fixed over time. Then, $\Theta_2$ does
not have to readjust to compensate for the change in the distribution
of $\vx$.

Fixed distribution of inputs to a sub-network would have positive
consequences for the layers {\em outside} the sub-network, as
well. Consider a layer with a sigmoid activation function $\vz =
g(W\vu+\vb)$ where $\vu$ is the layer input, the weight matrix $W$ and
bias vector $\vb$ are the layer parameters to be learned, and $g(x) =
\frac{1}{1+\exp(-x)}$. As $|x|$ increases, $g'(x)$ tends to zero. This
means that for all dimensions of $\vx=W\vu+\vb$ except those with
small absolute values, the gradient flowing down to $\vu$ will vanish
and the model will train slowly. However, since $\vx$ is affected by
$W, \vb$ and the parameters of all the layers below, changes to those
parameters during training will likely move many dimensions of $\vx$
into the saturated regime of the nonlinearity and slow down the
convergence. This effect is amplified as the network depth
increases. In practice, the saturation problem and the resulting
vanishing gradients are usually addressed by using Rectified Linear
Units \cite{relu} $ReLU(x)=\max(x,0)$, careful initialization
\cite{glorot-difficulty,iclr-dynamics}, and small learning rates.  If,
however, we could ensure that the distribution of nonlinearity inputs
remains more stable as the network trains, then the optimizer would be
less likely to get stuck in the saturated regime, and the training
would accelerate.

We refer to the change in the distributions of internal nodes of a
deep network, in the course of training, as {\em Internal Covariate Shift}. Eliminating it offers
a promise of faster training.  We propose a new mechanism, which we
call {\em Batch Normalization}, that takes a step towards reducing
internal covariate shift, and in doing so dramatically accelerates the
training of deep neural nets. It accomplishes this via a normalization
step that fixes the means and variances of layer inputs. Batch
Normalization also has a beneficial effect on the gradient flow
through the network, by reducing the dependence of gradients on the
scale of the parameters or of their initial values. This allows us to
use much higher learning rates without the risk of
divergence. Furthermore, batch normalization regularizes the model and
reduces the need for Dropout \cite{dropout}.  Finally, Batch
Normalization makes it possible to use saturating nonlinearities by
preventing the network from getting stuck in the saturated modes.

In Sec.~\ref{sec-results}, we apply Batch Normalization to the
best-performing ImageNet classification network, and show that we can
match its performance using only 7\% of the training steps, and can
further exceed its accuracy by a substantial margin.  Using an
ensemble of such networks trained with Batch Normalization, we achieve
the top-5 error rate that improves upon the best known results on
ImageNet classification.

\section{Towards Reducing Internal \mbox{Covariate} Shift}

We define {\em Internal Covariate Shift} as
the change in the
distribution of network activations due to the change in network parameters during training.  To improve the
training, we seek to reduce the internal covariate shift.  By fixing
the distribution of the layer inputs $\vx$ as the training progresses,
we expect to improve the training speed.
It has been long known \cite{lecun-backprop,
  loglinear-training}
that the network training converges faster if its inputs are whitened -- i.e.,
linearly transformed to have zero means and unit variances, and  decorrelated. As each layer
observes the inputs produced by the layers below, it would be advantageous to
achieve the same whitening of the inputs of each layer.  By whitening the
inputs to each layer, we would take a step towards achieving the fixed
distributions of inputs that would remove the ill effects of the internal covariate shift.

We could consider whitening activations at every training step or at
some interval, either by modifying the network directly or by changing
the parameters of the optimization algorithm to depend on the network
activation values \cite{mean-normalized-sgd, raiko, povey,
  desjardins}.  However, if these modifications are interspersed with
the optimization steps, then the gradient descent step may attempt to
update the parameters in a way that requires the normalization to be
updated, which reduces the effect of the gradient step. For example,
consider a layer with the input $u$ that adds the learned bias $b$,
and normalizes the result by subtracting the mean of the activation
computed over the training data: $\xhat=x - E[x]$ where $x = u+b$,
\,$\setX=\{x_{1\ldots N}\}$ is the set of values of $x$ over the
 training set, and $ \E[x] = \frac{1}{N}\sum_{i=1}^N
x_i$. If  a gradient descent step ignores the dependence of $\E[x]$  on $b$, then it  will
update $b\leftarrow b+\Delta b$, where $\Delta b\propto -\partial{\ell}/\partial{\xhat}$. Then  $u+(b+\Delta b) -
\E[u+(b+\Delta b)] =
u+b-\E[u+b]$. Thus, the combination of the update to $b$ and subsequent change in
normalization led to no change in the output of the layer nor,
consequently, the loss. As the training continues, $b$ will grow
indefinitely while the loss remains fixed. This problem can get worse
if the normalization not only centers but also scales the activations.
We have observed this empirically in initial experiments, where the
model blows up when the normalization parameters are computed outside
the gradient descent step.

The issue with the above approach is that the gradient descent
optimization does not take into account the fact that the
normalization takes place.  To address this issue, we would like to
ensure that, for any parameter values, the network {\em always}
produces activations with the desired distribution. Doing so would
allow the gradient of the loss with respect to the model parameters to
account for the normalization, and for its dependence on the model
parameters $\Theta$. Let again $\vx$ be a layer input, treated as a
vector, and $\setX$ be the set of these inputs over the training data
set. The normalization can then be written as a transformation
$$\vxhat=\Norm(\vx,\setX)$$ which depends not only on the given
training example $\vx$ but on all examples $\setX$ -- each of which
depends on $\Theta$ if $\vx$ is generated by another layer. For
backpropagation, we would need to compute the Jacobians
$$\jac{\Norm(\vx,\setX)}{\vx} \text{\, and\, }\jac{\Norm(\vx,\setX)}{\setX};$$
ignoring the latter term would lead to the explosion described above.
Within this framework, whitening the layer inputs is expensive, as it requires
computing the covariance matrix $\Cov[\vx]=\E_{\vx\in\setX}[\vx \vx^T]-\E[\vx]\E[\vx]^T$ and its
inverse square root, to produce the whitened activations $\Cov[\vx]^{-1/2}(\vx-\E[\vx])$,
as well as the derivatives of these transforms for backpropagation.
 This motivates us to seek an
alternative that performs input normalization in a way that is
differentiable and does not require the analysis of the entire
training set after every parameter update.


Some of the previous approaches
(e.g. \cite{lyu-simoncelli}) use statistics computed over a single
training example, or, in the case of image networks, over different
feature maps at a given location. However, this changes the
representation ability of a network by discarding the absolute scale
of activations. We want to a preserve the information in the network,
by normalizing the activations in a training example relative to the
statistics of the entire training data.

\section{Normalization via Mini-Batch Statistics}

Since the full whitening of each layer's inputs is costly and not
everywhere differentiable, we make two necessary simplifications. The first
 is that instead of whitening the features in layer
inputs and outputs jointly, we will normalize each scalar feature
independently, by making it have the mean of zero and the variance of
1. For a layer with $d$-dimensional input $\vx = (x^{(1)}\ldots x^{(d)})$, we
will normalize each dimension 
$$\xhat^\kk = \frac{x^\kk-\E[x^\kk]}{
  \sqrt{\Var[x^\kk]}}$$
where the expectation and variance are
computed over the training data set. As shown in
\cite{lecun-backprop}, such normalization speeds up convergence,
even when the  features are not decorrelated.

Note that simply normalizing each input of a layer may change what the
layer can represent. For instance, normalizing the inputs of a
sigmoid would constrain them to the linear
regime of the nonlinearity. To address this, we make sure that {\em the transformation inserted in
  the network can represent the identity transform}.  To
accomplish this, we introduce, for each activation $x^\kk$, a pair of
parameters $\gamma^\kk, \beta^\kk$, which scale and shift the
normalized value: $$y^\kk = \gamma^\kk\xhat^\kk +
\beta^\kk.$$ These parameters are learned along with the original model
parameters, and restore the representation power of the
network. Indeed, by setting $\gamma^\kk = \sqrt{\Var[x^\kk]}$ and
$\beta^\kk = \E[x^\kk]$, we could recover the original activations, if that were the optimal thing to do.

In the batch setting where each training step is based on the entire training
set, we would use the whole set to normalize activations. However, this is
impractical when using stochastic optimization. Therefore, we make the second
simplification: since we use mini-batches in stochastic gradient training, {\em
  each mini-batch produces estimates of the mean and variance} of each
activation. This way, the statistics used for normalization can fully
participate in the gradient backpropagation.
Note that the use of mini-batches is enabled by computation of
per-dimension variances rather than joint covariances; in the joint case,
regularization would be required since the mini-batch size is likely to be
smaller than the number of activations being whitened, resulting in singular
covariance matrices.

Consider a mini-batch $\setB$ of size $m$. Since the normalization is applied to
each activation independently, let us focus on a particular activation $x^\kk$ and omit $k$ for clarity. We have $m$ values of this activation
in the mini-batch,
$$\setB=\{x_{1\ldots m}\}.$$ Let the normalized values be
$\xhat_{1\ldots m}$, and their linear transformations be $y_{1\ldots m}$. We refer to the transform $$\BatchNorm_{\gamma,\beta}: x_{1\ldots m}\rightarrow y_{1\ldots m}$$ as the {\em Batch Normalizing  Transform}.
  We present the BN Transform in Algorithm~\ref{alg-bn}.  In the algorithm, $\epsilon$ is a constant  added to the mini-batch variance for numerical stability.

\begin{algorithm}
  \caption{Batch Normalizing Transform, applied to \mbox{activation $x$} over a mini-batch. }
\label{alg-bn}
  \begin{algorithmic}
  \REQUIRE 
  \begin{tabular}[t]{@{}l}Values of   $x$ over a mini-batch:
  $\setB=\{x_{1\ldots m}\}$;\\ 
 Parameters to be learned: $\gamma$,
    $\beta$ \end{tabular}
  \ENSURE $\{y_i =  \BN{x_i}{\gamma,\beta}\}$
  \begin{flalign*}
      \mu_\setB &\leftarrow \frac{1}{m}\sum_{i=1}^m x_i &\text{\comt mini-batch mean}&\\
  \sigma_\setB^2 &\leftarrow \frac{1}{m}\sum_{i=1}^m (x_i-\mu_\setB)^2& \text{\comt mini-batch variance}&\\
\xhat_i &\leftarrow \frac{x_i-\mu_\setB}{\sqrt{\sigma_\setB^2+\epsilon}}   
&\text{\comt normalize}&\\
  y_i &\leftarrow \gamma\xhat_i + \beta  
  \equiv\BN{x_i}{\gamma,\beta}
    &\text{\comt scale and shift}&
  \end{flalign*}
\end{algorithmic}
\end{algorithm}

The BN transform  can be added to a network to manipulate any activation. In the notation $y = \BN{x}{\gamma,\beta}$, we indicate that the parameters $\gamma$ and $\beta$ are to be learned, but it should be noted that the BN transform does not independently process the activation in each training example. Rather,  $\BN{x}{\gamma,\beta}$ depends both on the training example {\em and the other examples in the mini-batch}.
The scaled and shifted values $y$  are passed to other network layers. The normalized activations $\xhat$ are internal to our transformation, but their presence is crucial. The distributions of  values of any  $\xhat$ has the
expected value of $0$ and the variance of $1$, as long as the elements of each mini-batch are 
 sampled from the same distribution, and if we neglect $\epsilon$.  This can be seen by observing that $\sum_{i=1}^m \xhat_i = 0$ and
$\frac{1}{m}\sum_{i=1}^m \xhat_i^2 = 1$, and  taking expectations. Each normalized activation $\xhat^\kk$ can be viewed as an input to a sub-network composed of the linear transform $y^\kk=\gamma^\kk\xhat^\kk+\beta^\kk$, followed by the other processing done by the original network. These sub-network inputs all have fixed means and variances, and although the joint distribution of these normalized $\xhat^\kk$ can change over the course of  training, we expect that the introduction of  normalized inputs accelerates the training of the sub-network and, consequently, the network as a whole. 
  
During training we need to  backpropagate the gradient of loss $\ell$ through
this transformation, as well  as  compute the  gradients with respect to the parameters of the BN transform. We use  chain rule, as follows (before
simplification):
\begin{align*}
\textstyle\jac{\ell}{\xhat_i} &\textstyle= \jac{\ell}{y_i}\cdot \gamma \\ 
\textstyle\jac{\ell}{\sigma_\setB^2}
&\textstyle= \sum_{i=1}^m \jac{\ell}{\xhat_i}\cdot(x_i-\mu_\setB)\cdot
\frac{-1}{2}(\sigma_\setB^2+\epsilon)^{-3/2} \\ 
\textstyle\jac{\ell}{\mu_\setB} &\textstyle=
\bigg(\sum_{i=1}^m \jac{\ell}{\xhat_i}\cdot
\frac{-1}{\sqrt{\sigma_\setB^2+\epsilon}}\bigg) +
\jac{\ell}{\sigma_\setB^2}\cdot\frac{   \sum_{i=1}^m
  -2(x_i-\mu_\setB)}{m}\\
 \textstyle  \jac{\ell}{x_i} &\textstyle= \jac{\ell}{\xhat_i} \cdot
\frac{1}{\sqrt{\sigma_\setB^2+\epsilon}} + \jac{\ell}{\sigma_\setB^2}\cdot
\frac{2(x_i-\mu_\setB)}{m} + \jac{\ell}{\mu_\setB}\cdot \frac{1}{m}\\
\textstyle\jac{\ell}{\gamma}&\textstyle= \sum_{i=1}^m \jac{\ell}{y_i} \cdot \xhat_i
  \\ 
\textstyle  \jac{\ell}{\beta} &\textstyle= \sum_{i=1}^m \jac{\ell}{y_i}
\end{align*}
Thus, BN transform is a differentiable transformation that introduces  normalized activations
into the network. This ensures that as the model is training, layers can continue learning on input distributions that exhibit less internal covariate shift, thus accelerating the training.
Furthermore, the learned affine transform applied to these normalized activations allows the BN transform  to represent the identity transformation and preserves the network capacity.

\subsection{Training and Inference with Batch-Normalized Networks}
\label{sec-training}

To {\em Batch-Normalize} a network, we specify a subset of activations
and insert the BN transform for each of them, according to
Alg.~\ref{alg-bn}. Any layer that previously received $x$ as the
input, now receives $\BatchNorm(x)$.  A model employing Batch
Normalization can be trained using batch gradient descent, or
Stochastic Gradient Descent with a mini-batch size $m>1$, or with any
of its variants such as Adagrad \cite{adagrad}.  The normalization of activations
that depends on the mini-batch allows efficient training, but is
neither necessary nor desirable during inference; we want the output
to depend only on the input, deterministically. For this, once the
network has been trained, we use the
normalization $$\xhat=\frac{x-\E[x]}{\sqrt{\Var[x]+\epsilon}}$$ using
the population, rather than mini-batch, statistics. Neglecting
$\epsilon$, these normalized activations have the same mean 0 and
variance 1 as during training. We use the unbiased variance estimate
$\Var[x] = \frac{m}{m-1}\cdot\E_\setB[\sigma_\setB^2]$, where the
expectation is over training mini-batches of size $m$ and
$\sigma_\setB^2$ are their sample variances. Using moving averages
instead, we can track the accuracy of a model as it trains.  Since the
means and variances are fixed during inference, the normalization is
simply a linear transform applied to each activation. It may further
be composed with the scaling by $\gamma$ and shift by $\beta$, to
yield a single linear transform that replaces $\BatchNorm(x)$.
Algorithm~\ref{alg-train} summarizes the procedure for training
batch-normalized networks.

\begin{algorithm}
\caption{Training a Batch-Normalized Network}
\label{alg-train}
\begin{algorithmic}[1]
\REQUIRE 
\begin{tabular}[t]{@{}l}
Network $\norig$ with trainable  parameters $\Theta$;\\
 subset of activations $\{x^\kk\}_{k=1}^K$
 \end{tabular}
\ENSURE  Batch-normalized network  for inference, $\ninf$
\STATE $\ntrain\leftarrow \norig$ \quad \COMMENT Training BN network
\FOR{$k = 1\ldots K$}
\STATE 
Add transformation $y^\kk = \BN{x^\kk}{\gamma^\kk,\beta^\kk}$  to $\ntrain$ (Alg.~\ref{alg-bn})
\STATE
Modify each layer in $\ntrain$ with input $x^\kk$ to take $y^\kk$ instead
\ENDFOR
\STATE
Train $\ntrain$ to optimize the parameters $\Theta\cup 
\{\gamma^\kk, \beta^\kk\}_{k=1}^K$
\STATE \vspace{.03in}
$\ninf\leftarrow\ntrain$\quad \begin{tabular}[t]{@{}l}\COMMENT Inference BN network with frozen\\ \COMMENT parameters \end{tabular}
\FOR{$k = 1\ldots K$}
\STATE \COMMENT{For clarity, $x\equiv x^\kk, \gamma\equiv\gamma^\kk, \mu_\setB\equiv\mu_\setB^\kk$, etc.}
\STATE Process multiple training mini-batches  $\setB$, each of size $m$, and average over them:
\vspace{-.1in}
\begin{align*}
\E[x] &\leftarrow \E_\setB[\mu_\setB]\\
 \Var[x] &\leftarrow \textstyle \frac{m}{m-1}\E_\setB[\sigma_\setB^2]
 \end{align*}
\STATE 
\vspace{-.1in}
In $\ninf$, replace the transform $y=\BN{x}{\gamma,\beta}$ with\, $y = \frac{\gamma}{\sqrt{\Var[x]+\epsilon}}\cdot x + \big(\beta - \frac{\gamma\,\E[x]}{\sqrt{\Var[x]+\epsilon}}\big)$
\ENDFOR
\end{algorithmic}
\end{algorithm}

\subsection{Batch-Normalized Convolutional Networks}
\label{sec-conv}

Batch Normalization can be applied to any set of activations in the network. Here, we focus on transforms that
consist of an affine transformation  followed by an element-wise
nonlinearity: $$\vz = g(W\vu+\vb)$$ where $W$ and $\vb$ are learned parameters of the
model, and $g(\cdot)$ is the nonlinearity such as sigmoid or
ReLU. This formulation covers both fully-connected and convolutional layers. We add the BN transform immediately before the nonlinearity, by normalizing $\vx=W\vu+\vb$.  We could have also normalized the layer inputs $\vu$, but 
since $\vu$ is likely the output of another nonlinearity, the
shape of its distribution is likely to change during training, and constraining its first and second moments would not eliminate the covariate shift.
In contrast,  $W\vu+\vb$ is more likely to have a symmetric, non-sparse distribution,
that is ``more Gaussian'' \cite{ica}; normalizing it is likely to produce activations with a stable  distribution.

Note that, since we normalize $W\vu+\vb$, the bias $\vb$ can be ignored since its
effect will be canceled by the subsequent mean subtraction (the role of the bias is subsumed by $\beta$ in Alg.~\ref{alg-bn}). Thus,  $\vz = g(W\vu+\vb)$  is replaced with
$$
\vz = g(\BatchNorm(W\vu))
$$
where the BN transform is applied independently to each dimension of $\vx=W\vu$, with a separate pair of learned parameters $\gamma^\kk$, $\beta^\kk$ per dimension.

For convolutional layers, we  additionally want the normalization
to  obey the convolutional property -- so that different elements
of the same feature map, at different locations, are normalized in the
same way. To achieve this, we jointly normalize all the activations in
a mini-batch, over all locations. In Alg.~\ref{alg-bn}, we let
$\setB$ be the set of all values in a feature map across both the
elements of a mini-batch and spatial locations -- so for a mini-batch
of size $m$ and feature maps of size $p\times q$, we use the effective mini-batch of size $m'=|\setB| =
m\cdot p\, q$. We learn a pair of parameters $\gamma^\kk$ and $\beta^\kk$ per feature map, rather than per activation.
Alg.~\ref{alg-train} is modified similarly, so that during inference the BN transform applies the same linear transformation to each activation in a given feature map. 
	
\subsection{ Batch Normalization enables higher learning rates} 
\label{sec-lr}
 In
traditional deep networks, too-high learning rate may result in the gradients that explode or vanish, as well as getting stuck in poor local minima. Batch Normalization helps address these issues. By normalizing activations throughout the network, it prevents small changes to the parameters from amplifying into larger and suboptimal changes in activations in gradients; for instance, it prevents the training from getting stuck in the saturated regimes of nonlinearities.   

Batch Normalization also makes training more resilient to the parameter scale. Normally, large learning rates 
may increase the scale of layer parameters, which then amplify the gradient during backpropagation and lead to the model explosion.
  However, with Batch Normalization, backpropagation through a layer is unaffected by the scale of its parameters.  Indeed, for a scalar $a$,  $$\BatchNorm(W\vu) =
\BatchNorm((aW)\vu)$$ and we can show that
\begin{align*}
\textstyle\jac{\BatchNorm((aW)\vu)}{\vu}&= \textstyle
\jac{\BatchNorm(W\vu)}{\vu} \\
\textstyle\jac{\BatchNorm((aW)\vu)}{(aW)}&\textstyle =\frac{1}{a}\cdot
\jac{\BatchNorm(W\vu)}{W}
\end{align*}
The scale does not affect the layer Jacobian nor, consequently, the gradient propagation. Moreover, larger weights lead to {\em smaller} gradients, and Batch Normalization will stabilize the parameter growth.

We further conjecture that Batch Normalization may lead the layer Jacobians to
have singular values close to 1, which is known to be beneficial for training \cite{iclr-dynamics}. Consider two consecutive layers with normalized inputs, and the transformation  between these  normalized  vectors:
$\vzhat = F(\vxhat)$. If we assume that  $\vxhat$ and $\vzhat$ are Gaussian and uncorrelated, and 
that $F(\vxhat)\approx J\vxhat$ is a linear transformation for the given model
parameters, then both $\vxhat$ and $\vzhat$ have unit covariances, and  $I=\Cov[\vzhat] =J \Cov[\vxhat] J^T = JJ^T$. Thus, $JJ^T=I$, and so all
singular values of $J$ are equal to 1, which  preserves the
gradient magnitudes during backpropagation. In reality, the transformation is
not linear, and the normalized values are not guaranteed to be Gaussian nor independent, but we nevertheless expect Batch Normalization to help make gradient propagation  better behaved. The precise effect of Batch
Normalization on gradient propagation remains an area of further study.

\subsection{Batch Normalization  regularizes the model} 
\label{sec-regularizer}
When training with Batch Normalization, a training example is seen in
conjunction with other examples in the mini-batch, and the training network no longer
producing deterministic values for a given training example. In our
experiments, we found  this effect to be advantageous to the
generalization of the network. Whereas Dropout \cite{dropout} is
typically used to reduce overfitting, in a batch-normalized network
we found that it can be either removed  or reduced in strength.

\section{Experiments}

\subsection{Activations over time}

To verify the effects of internal covariate shift on training, and the ability of Batch
Normalization to combat it, we considered the problem of predicting the digit
class on the MNIST dataset \cite{mnist}. We used a very simple network, with  a 28x28
binary image as  input, and  3 fully-connected hidden layers with 100 activations each. 
 Each hidden layer computes $\vy = g(W\vu+\vb)$ with  sigmoid nonlinearity, and the weights $W$ initialized to small random  Gaussian values. The last hidden layer is followed by a fully-connected layer with 10 activations (one per class) and  cross-entropy loss. We trained the network for 50000
steps, with  60 examples per mini-batch. We added Batch Normalization to each hidden layer of the network, as in Sec.~\ref{sec-training}.
 We were interested in the
comparison between the baseline and batch-normalized networks, rather than
achieving the state of the art performance on MNIST (which the described
architecture does not).

\begin{figure}
\centering
\begin{tabular}{@{}c@{\,}c@{}c@{}}
\includegraphics[width=0.28\columnwidth]{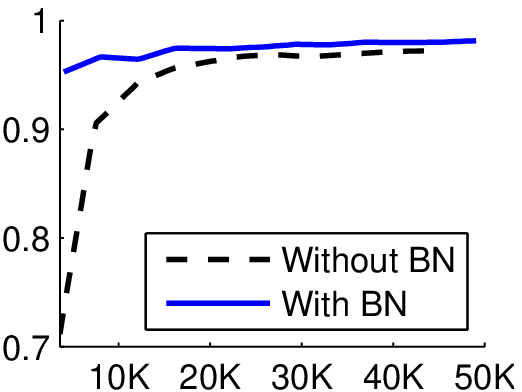}
&
\includegraphics[width=0.35\columnwidth]{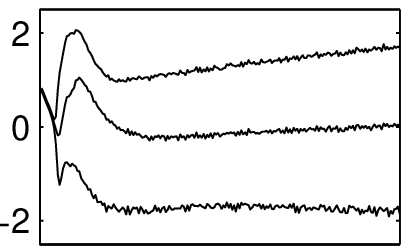}&
\includegraphics[width=0.35\columnwidth]{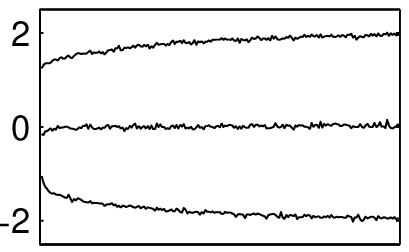}\\
(a)&(b) Without BN&(c) With BN
\end{tabular} 
\caption{\em {\em(a)} The test accuracy of the MNIST network trained with and without Batch Normalization, vs. the number of training steps. Batch Normalization helps the network train faster and achieve higher accuracy. {\em(b, c)}   The evolution of input distributions to a typical sigmoid, over the course of training, shown as $\{15, 50, 85\}$th percentiles. Batch Normalization makes the distribution more stable and reduces the internal covariate shift.}
\label{fig-mnist}
\end{figure}

Figure~\ref{fig-mnist}(a) shows the fraction of correct predictions  by the two networks on
held-out test data, as training progresses. The batch-normalized network enjoys the higher test accuracy. To investigate why, we studied inputs to the sigmoid, in the original network $\norig$ and  batch-normalized network $\ntrain$ (Alg.~\ref{alg-train}) over the course of training. In Fig.~\ref{fig-mnist}(b,c) we show, for one typical activation from the last hidden layer of each network, how its distribution evolves.  The distributions in the original network
change significantly over time, both in their mean and the variance, which complicates the training of the subsequent layers.  In
contrast, the distributions in the batch-normalized network are much more stable
as training progresses, which aids the training.

\subsection{ImageNet classification}
\label{sec-results}

We applied Batch Normalization to a new variant of the Inception network \cite{inception},
trained on the ImageNet classification task \cite{imagenet}. The network has a large
number of convolutional and pooling layers, with a softmax layer to predict the
image class, out of 1000 possibilities. Convolutional layers use ReLU as the
nonlinearity. The main difference to the network described in \cite{inception} is that
the $5\times 5$ convolutional layers are replaced by two consecutive layers of $3\times 3$ convolutions
with up to $128$ filters. The network contains $13.6\mils$ parameters, and, other than the top softmax layer, has no fully-connected layers.  More details are given in the Appendix.  We refer to this model as {\sl Inception} in the rest of the text. The model was trained using a version of Stochastic Gradient Descent with momentum
\cite{momentum}, using the mini-batch size of 32. The training was performed using a large-scale, distributed architecture (similar to \cite{dist-belief}).
All networks are evaluated as training progresses by computing the validation accuracy $@1$, i.e. the
probability of predicting the correct label out of 1000 possibilities, on a held-out set, using a single crop per image.

\begin{figure*}
\centering
\begin{minipage}[b]{\columnwidth}
\begin{tabular}{@{}c@{}}
\includegraphics[width=\columnwidth]{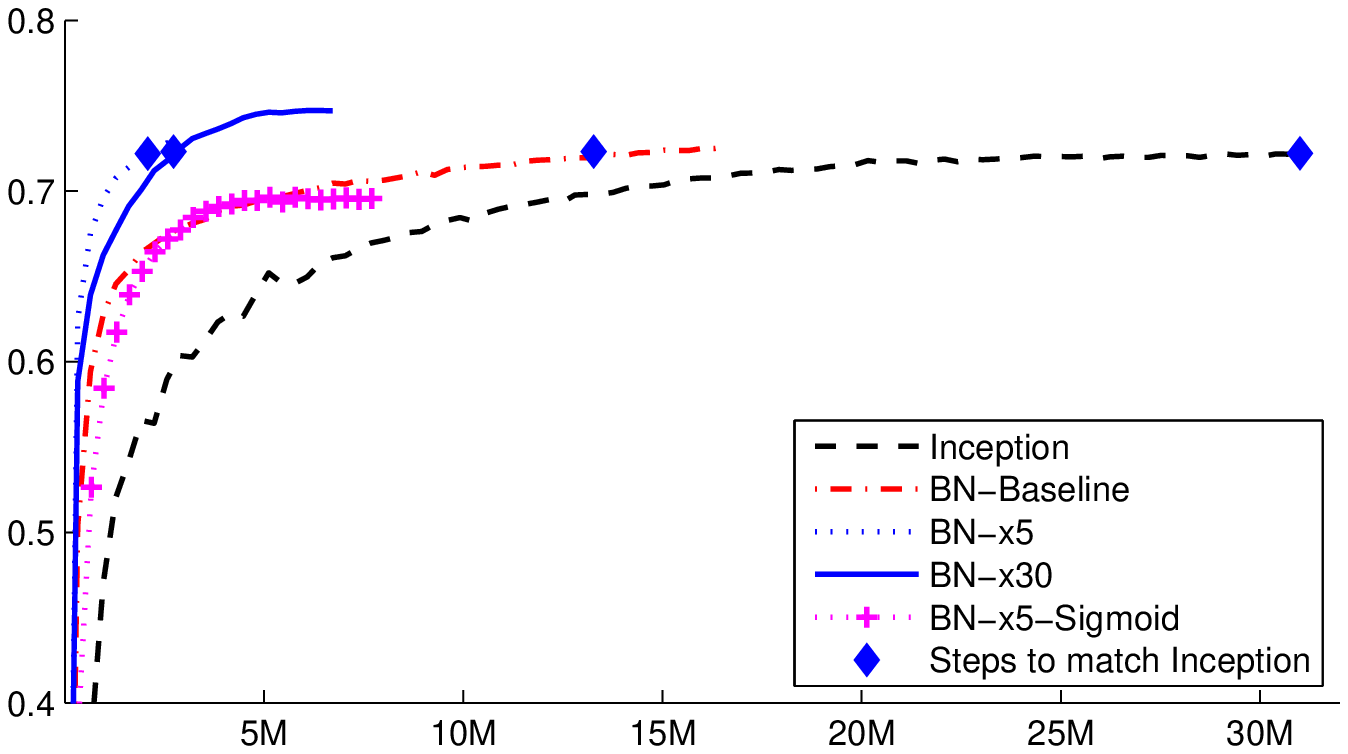}
\end{tabular} 
\caption{\em Single crop validation accuracy of Inception and its
  batch-normalized variants, vs. the number of training steps.  }
\label{fig-inception}
\end{minipage}
\qquad
\begin{minipage}[b]{0.9\columnwidth}
\begin{tabular}{@{} c | r  r  @{}}
\hline
Model & Steps to  72.2\% & Max accuracy \\ 
\hline
Inception& $31.0\mils$ & 72.2\%  \\
\sl BN-Baseline& $13.3\mils$ & 72.7\%  \\
\sl BN-x5& $2.1\mils$ & 73.0\%  \\
\sl BN-x30& $2.7\mils$ & 74.8\% \\
\sl BN-x5-Sigmoid&  & 69.8\%\\\hline
\end{tabular}
\caption{\em For Inception and the batch-normalized variants, the number of training steps required to reach the maximum accuracy of Inception (72.2\%), and the maximum accuracy achieved by the network.}
\label{fig-stats}
\end{minipage}
\end{figure*}

In our experiments, we evaluated several modifications of Inception with Batch Normalization. In all cases, Batch Normalization was applied to the 
input of each nonlinearity, in a convolutional way, as described in section
\ref{sec-conv}, while keeping the rest of the architecture constant.

\subsubsection{Accelerating BN Networks}
\label{sec-accelerating}
Simply adding Batch Normalization to a network does not take full advantage of our method. To do so, we further changed the network and its training parameters, as follows:

{\em Increase learning rate.} In a batch-normalized model, we have been able to achieve a training speedup from higher learning rates, with no ill side effects (Sec.~\ref{sec-lr}).

 {\em Remove Dropout.} As described in Sec.~\ref{sec-regularizer}, Batch Normalization fulfills some of the same goals as Dropout. Removing Dropout from Modified BN-Inception speeds up training, without increasing overfitting. 
 
{\em Reduce the $L_2$ weight regularization.} While in Inception an $L_2$ loss on the model parameters controls overfitting, in Modified BN-Inception the weight of this loss is reduced by a factor of 5. We find that this improves the accuracy on the held-out validation data.

{\em Accelerate the learning rate decay.} In training Inception, learning rate was decayed exponentially. Because our network trains faster than Inception, we lower the learning rate  6 times faster.

{\em Remove Local Response Normalization}   While Inception and other networks \cite{dropout} benefit from it, we found that with Batch Normalization  it is not necessary.

{\em Shuffle training examples more thoroughly.} We enabled within-shard shuffling of the training data, which prevents the same examples from always appearing in a mini-batch together. This  led to about 1\% improvements in the validation accuracy, which is consistent with the view of Batch Normalization as a regularizer (Sec.~\ref{sec-regularizer}): the randomization inherent in our method should be most beneficial when it  affects an example differently each time it is seen.

{\em Reduce the photometric distortions.} Because batch-normalized networks train faster and observe each training example fewer times, we let the trainer focus on more ``real'' images by distorting them less.

\subsubsection{Single-Network Classification}

We evaluated the following networks, all trained on the LSVRC2012 training data, and tested on the validation data:

\netw{Inception}: the network described at the beginning of Section \ref{sec-results}, trained with the initial learning rate of 0.0015.

\netw{BN-Baseline}: Same as Inception with Batch Normalization before each nonlinearity.

\netw{BN-x5}: Inception with Batch Normalization and the modifications  in Sec.~\ref{sec-accelerating}. The initial learning rate was increased by a factor of 5, to 0.0075. The same learning rate increase with original Inception caused the model parameters to reach machine infinity.

\netw{BN-x30}: Like \netw{BN-x5}, but with the initial learning rate  0.045 (30 times that of Inception).

\netw{BN-x5-Sigmoid}: Like \netw{BN-x5}, but with sigmoid nonlinearity $g(t)=\frac{1}{1+\exp(-x)}$ instead of ReLU.
We also attempted to train the original Inception with sigmoid, but the model remained at the  accuracy equivalent to chance.

In Figure~\ref{fig-inception}, we show the validation accuracy of the
networks, as a function of the number of training steps.  Inception
reached the accuracy of 72.2\% after $31\mils$ training steps. The
Figure~\ref{fig-stats} shows, for each network, the number of training
steps required to reach the same 72.2\% accuracy, as well as the
maximum validation accuracy reached by the network and the number of steps
to reach it.
 
By only using Batch Normalization (\netw{BN-Baseline}), we match the accuracy of Inception in less than half the number of training steps. By applying the modifications in Sec.~\ref{sec-accelerating}, we significantly increase the training speed of the network. \netw{BN-x5} needs 14 times fewer steps than Inception to reach the 72.2\% accuracy.
Interestingly, increasing the learning rate further (\netw{BN-x30})  causes the model to train somewhat {\em slower} initially, but allows it to reach a higher final accuracy. It  reaches  74.8\% after $6\mils$ steps, i.e. 5 times fewer steps than required by Inception to reach 72.2\%.

We also verified that the  reduction in internal covariate shift allows deep networks with Batch Normalization to be trained when sigmoid is used as the nonlinearity, despite the well-known difficulty of training such networks. Indeed, \netw{BN-x5-Sigmoid} achieves the accuracy of  69.8\%. Without Batch Normalization, Inception with sigmoid never achieves better than $1/1000$ accuracy.

\subsubsection{Ensemble Classification}

\begin{figure*}[t!]
\centering
\begin{tabular}{c | r  r  r  r  r }
\hline
Model & Resolution & Crops & Models & Top-1 error & Top-5 error \\ 
\hline
{GoogLeNet ensemble} & 224 & 144 & 7 & - & 6.67\% \\
{Deep Image low-res} & 256 & - & 1 & - & 7.96\% \\
{Deep Image high-res} & 512 & - & 1 & 24.88 & 7.42\% \\
{Deep Image ensemble} & variable & - & - & - & 5.98\% \\
{BN-Inception single crop} & 224 & 1 & 1 & 25.2\% & 7.82\% \\
{BN-Inception multicrop} & 224 & 144 & 1 & 21.99\% & 5.82\% \\
{BN-Inception ensemble} & 224 & 144 & 6 & 20.1\% & {\bf 4.9\%}* \\
\hline
\end{tabular}
\caption{\em Batch-Normalized Inception comparison with previous state of the art on the provided validation set comprising 50000 images.
  *BN-Inception ensemble has reached 4.82\% top-5 error on the 100000 images of the test set of the ImageNet as reported by the test server. }
\label{fig-classification-comparison}
\end{figure*}

The current reported best results on the ImageNet Large Scale Visual Recognition
Competition are reached by the Deep Image ensemble of traditional models
\cite{deepimage} and the ensemble model of \cite{msr}. The latter reports the
top-5 error of 4.94\%, as evaluated by the ILSVRC server. Here we report a top-5
validation error of 4.9\%, and test error of 4.82\% (according to the ILSVRC
server). This improves upon the previous best result,
and
exceeds the estimated accuracy of human raters according to \cite{imagenet}.


For our ensemble, we used 6 networks. Each was based on {\sl BN-x30},
modified via some of the following: increased initial weights in the
convolutional layers; using Dropout (with the Dropout probability of
5\% or 10\%, vs. 40\% for the original Inception); and using
non-convolutional, per-activation Batch Normalization with last hidden
layers of the model. Each network achieved its maximum accuracy after
about $6\mils$ training steps.  The ensemble prediction was based on
the arithmetic average of class probabilities predicted by the
constituent networks. The details of ensemble and multicrop inference
are similar to \cite{inception}.

We demonstrate in Fig.~\ref{fig-classification-comparison} that batch
normalization allows us to set new state-of-the-art by a healthy
margin on the ImageNet classification challenge benchmarks.

\section{Conclusion}

We have presented a novel mechanism for dramatically accelerating the
training of deep networks. It is based on the premise that covariate
shift, which is known to complicate the training of machine learning
systems, also applies to sub-networks and layers, and removing it from
internal activations of the network may aid in training. Our proposed
method draws its power from normalizing activations, and from
incorporating this normalization in the network architecture
itself. This ensures that the normalization is appropriately handled
by any optimization method that is being used to train the network. To
enable stochastic optimization methods commonly used in deep network
training, we perform the normalization for each mini-batch, and
backpropagate the gradients through the normalization
parameters. Batch Normalization adds only two extra parameters per
activation, and in doing so preserves the representation ability of
the network. We presented an algorithm for constructing, training, and
performing inference with batch-normalized networks. The resulting
networks can be trained with saturating nonlinearities, are more
tolerant to increased training rates, and often do not require Dropout
for regularization.

Merely adding Batch Normalization to a state-of-the-art image
classification model yields a substantial speedup in training. By
further increasing the learning rates, removing Dropout, and applying
other modifications afforded by Batch Normalization, we reach the
previous state of the art with only a small fraction of training steps
-- and then beat the state of the art in single-network image
classification. Furthermore, by combining multiple models trained with
Batch Normalization, we perform better than the best known system on
ImageNet, by a significant margin.

Interestingly, our method bears similarity to the standardization layer of
\cite{gulcehre}, though the two methods stem from very different goals, and
perform different tasks. The goal of Batch Normalization is to achieve a stable
distribution of activation values throughout training, and in our experiments we
apply it before the nonlinearity since that is where matching the first and
second moments is more likely to result in a stable distribution. On the
contrary, \cite{gulcehre} apply the standardization layer to the {\em output} of
the nonlinearity, which results in sparser activations. In our large-scale image
classification experiments, we have not observed the nonlinearity {\em inputs}
to be sparse, neither with nor without Batch Normalization. Other notable
differentiating characteristics of Batch Normalization include the learned scale
and shift that allow the BN transform to represent identity (the standardization
layer did not require this since it was followed by the learned linear transform
that, conceptually, absorbs the necessary scale and shift), handling of
convolutional layers, deterministic inference that does not depend on the
mini-batch, and batch-normalizing each convolutional layer in the network.

In this work, we have not explored the full range of possibilities
that Batch Normalization potentially enables. Our future work includes
applications of our method to Recurrent Neural Networks
\cite{pascanu-rnn}, where the internal covariate shift and the
vanishing or exploding gradients may be especially severe, and which
would allow us to more thoroughly test the hypothesis that
normalization improves gradient propagation (Sec.~\ref{sec-lr}). We
plan to investigate whether Batch Normalization can help with domain
adaptation, in its traditional sense -- i.e. whether the normalization
performed by the network would allow it to more easily generalize to
new data distributions, perhaps with just a recomputation of the
population means and variances (Alg.~\ref{alg-train}). Finally, we
believe that further theoretical analysis of the algorithm would allow
still more improvements and applications.

\bibliography{bnicml}

\begin{thebibliography}{24}
\providecommand{\natexlab}[1]{#1}
\providecommand{\url}[1]{\texttt{#1}}
\expandafter\ifx\csname urlstyle\endcsname\relax
  \providecommand{\doi}[1]{doi: #1}\else
  \providecommand{\doi}{doi: \begingroup \urlstyle{rm}\Url}\fi

\bibitem[Bengio \& Glorot(2010)Bengio and Glorot]{glorot-difficulty}
Bengio, Yoshua and Glorot, Xavier.
\newblock Understanding the difficulty of training deep feedforward neural
  networks.
\newblock In \emph{Proceedings of AISTATS 2010}, volume~9, pp.\  249--256, May
  2010.

\bibitem[Dean et~al.(2012)Dean, Corrado, Monga, Chen, Devin, Le, Mao, Ranzato,
  Senior, Tucker, Yang, and Ng]{dist-belief}
Dean, Jeffrey, Corrado, Greg~S., Monga, Rajat, Chen, Kai, Devin, Matthieu, Le,
  Quoc~V., Mao, Mark~Z., Ranzato, Marc'Aurelio, Senior, Andrew, Tucker, Paul,
  Yang, Ke, and Ng, Andrew~Y.
\newblock Large scale distributed deep networks.
\newblock In \emph{NIPS}, 2012.

\bibitem[Desjardins \& Kavukcuoglu()Desjardins and Kavukcuoglu]{desjardins}
Desjardins, Guillaume and Kavukcuoglu, Koray.
\newblock Natural neural networks.
\newblock (unpublished).

\bibitem[Duchi et~al.(2011)Duchi, Hazan, and Singer]{adagrad}
Duchi, John, Hazan, Elad, and Singer, Yoram.
\newblock Adaptive subgradient methods for online learning and stochastic
  optimization.
\newblock \emph{J. Mach. Learn. Res.}, 12:\penalty0 2121--2159, July 2011.
\newblock ISSN 1532-4435.

\bibitem[G{\"{u}}l{\c{c}}ehre \& Bengio(2013)G{\"{u}}l{\c{c}}ehre and
  Bengio]{gulcehre}
G{\"{u}}l{\c{c}}ehre, {\c{C}}aglar and Bengio, Yoshua.
\newblock Knowledge matters: Importance of prior information for optimization.
\newblock \emph{CoRR}, abs/1301.4083, 2013.

\bibitem[{He} et~al.(2015){He}, {Zhang}, {Ren}, and {Sun}]{msr}
{He}, K., {Zhang}, X., {Ren}, S., and {Sun}, J.
\newblock {Delving Deep into Rectifiers: Surpassing Human-Level Performance on
  ImageNet Classification}.
\newblock \emph{ArXiv e-prints}, February 2015.

\bibitem[Hyv\"{a}rinen \& Oja(2000)Hyv\"{a}rinen and Oja]{ica}
Hyv\"{a}rinen, A. and Oja, E.
\newblock Independent component analysis: Algorithms and applications.
\newblock \emph{Neural Netw.}, 13\penalty0 (4-5):\penalty0 411--430, May 2000.

\bibitem[Jiang(2008)]{domain-adaptation-survey}
Jiang, Jing.
\newblock A literature survey on domain adaptation of statistical classifiers,
  2008.

\bibitem[LeCun et~al.(1998{\natexlab{a}})LeCun, Bottou, Bengio, and
  Haffner]{mnist}
LeCun, Y., Bottou, L., Bengio, Y., and Haffner, P.
\newblock Gradient-based learning applied to document recognition.
\newblock \emph{Proceedings of the IEEE}, 86\penalty0 (11):\penalty0
  2278--2324, November 1998{\natexlab{a}}.

\bibitem[LeCun et~al.(1998{\natexlab{b}})LeCun, Bottou, Orr, and
  Muller]{lecun-backprop}
LeCun, Y., Bottou, L., Orr, G., and Muller, K.
\newblock Efficient backprop.
\newblock In Orr, G. and K., Muller (eds.), \emph{Neural Networks: Tricks of
  the trade}. Springer, 1998{\natexlab{b}}.

\bibitem[Lyu \& Simoncelli(2008)Lyu and Simoncelli]{lyu-simoncelli}
Lyu, S and Simoncelli, E~P.
\newblock Nonlinear image representation using divisive normalization.
\newblock In \emph{Proc. Computer Vision and Pattern Recognition}, pp.\  1--8.
  IEEE Computer Society, Jun 23-28 2008.
\newblock \doi{10.1109/CVPR.2008.4587821}.

\bibitem[Nair \& Hinton(2010)Nair and Hinton]{relu}
Nair, Vinod and Hinton, Geoffrey~E.
\newblock Rectified linear units improve restricted boltzmann machines.
\newblock In \emph{ICML}, pp.\  807--814. Omnipress, 2010.

\bibitem[Pascanu et~al.(2013)Pascanu, Mikolov, and Bengio]{pascanu-rnn}
Pascanu, Razvan, Mikolov, Tomas, and Bengio, Yoshua.
\newblock On the difficulty of training recurrent neural networks.
\newblock In \emph{Proceedings of the 30th International Conference on Machine
  Learning, {ICML} 2013, Atlanta, GA, USA, 16-21 June 2013}, pp.\  1310--1318,
  2013.

\bibitem[Povey et~al.(2014)Povey, Zhang, and Khudanpur]{povey}
Povey, Daniel, Zhang, Xiaohui, and Khudanpur, Sanjeev.
\newblock Parallel training of deep neural networks with natural gradient and
  parameter averaging.
\newblock \emph{CoRR}, abs/1410.7455, 2014.

\bibitem[Raiko et~al.(2012)Raiko, Valpola, and LeCun]{raiko}
Raiko, Tapani, Valpola, Harri, and LeCun, Yann.
\newblock Deep learning made easier by linear transformations in perceptrons.
\newblock In \emph{International Conference on Artificial Intelligence and
  Statistics ({AISTATS})}, pp.\  924--932, 2012.

\bibitem[Russakovsky et~al.(2014)Russakovsky, Deng, Su, Krause, Satheesh, Ma,
  Huang, Karpathy, Khosla, Bernstein, Berg, and Fei-Fei]{imagenet}
Russakovsky, Olga, Deng, Jia, Su, Hao, Krause, Jonathan, Satheesh, Sanjeev, Ma,
  Sean, Huang, Zhiheng, Karpathy, Andrej, Khosla, Aditya, Bernstein, Michael,
  Berg, Alexander~C., and Fei-Fei, Li.
\newblock {ImageNet Large Scale Visual Recognition Challenge}, 2014.

\bibitem[Saxe et~al.(2013)Saxe, McClelland, and Ganguli]{iclr-dynamics}
Saxe, Andrew~M., McClelland, James~L., and Ganguli, Surya.
\newblock Exact solutions to the nonlinear dynamics of learning in deep linear
  neural networks.
\newblock \emph{CoRR}, abs/1312.6120, 2013.

\bibitem[Shimodaira(2000)]{covariate-shift}
Shimodaira, Hidetoshi.
\newblock Improving predictive inference under covariate shift by weighting the
  log-likelihood function.
\newblock \emph{Journal of Statistical Planning and Inference}, 90\penalty0
  (2):\penalty0 227--244, October 2000.

\bibitem[Srivastava et~al.(2014)Srivastava, Hinton, Krizhevsky, Sutskever, and
  Salakhutdinov]{dropout}
Srivastava, Nitish, Hinton, Geoffrey, Krizhevsky, Alex, Sutskever, Ilya, and
  Salakhutdinov, Ruslan.
\newblock Dropout: A simple way to prevent neural networks from overfitting.
\newblock \emph{J. Mach. Learn. Res.}, 15\penalty0 (1):\penalty0 1929--1958,
  January 2014.

\bibitem[Sutskever et~al.(2013)Sutskever, Martens, Dahl, and Hinton]{momentum}
Sutskever, Ilya, Martens, James, Dahl, George~E., and Hinton, Geoffrey~E.
\newblock On the importance of initialization and momentum in deep learning.
\newblock In \emph{ICML (3)}, volume~28 of \emph{JMLR Proceedings}, pp.\
  1139--1147. JMLR.org, 2013.

\bibitem[Szegedy et~al.(2014)Szegedy, Liu, Jia, Sermanet, Reed, Anguelov,
  Erhan, Vanhoucke, and Rabinovich]{inception}
Szegedy, Christian, Liu, Wei, Jia, Yangqing, Sermanet, Pierre, Reed, Scott,
  Anguelov, Dragomir, Erhan, Dumitru, Vanhoucke, Vincent, and Rabinovich,
  Andrew.
\newblock Going deeper with convolutions.
\newblock \emph{CoRR}, abs/1409.4842, 2014.

\bibitem[Wiesler \& Ney(2011)Wiesler and Ney]{loglinear-training}
Wiesler, Simon and Ney, Hermann.
\newblock A convergence analysis of log-linear training.
\newblock In Shawe-Taylor, J., Zemel, R.S., Bartlett, P., Pereira, F.C.N., and
  Weinberger, K.Q. (eds.), \emph{Advances in Neural Information Processing
  Systems 24}, pp.\  657--665, Granada, Spain, December 2011.

\bibitem[Wiesler et~al.(2014)Wiesler, Richard, Schl{\"u}ter, and
  Ney]{mean-normalized-sgd}
Wiesler, Simon, Richard, Alexander, Schl{\"u}ter, Ralf, and Ney, Hermann.
\newblock Mean-normalized stochastic gradient for large-scale deep learning.
\newblock In \emph{IEEE International Conference on Acoustics, Speech, and
  Signal Processing}, pp.\  180--184, Florence, Italy, May 2014.

\bibitem[Wu et~al.(2015)Wu, Yan, Shan, Dang, and Sun]{deepimage}
Wu, Ren, Yan, Shengen, Shan, Yi, Dang, Qingqing, and Sun, Gang.
\newblock Deep image: Scaling up image recognition, 2015.

\end{thebibliography}
\bibliographystyle{icml2015}

\section*{Appendix}
\subsection*{Variant of the Inception Model Used}
\begin{figure*}[b]
{\small
\begin{center}
  \begin{tabular}[H]{@{}|l|c|c|c|c|c|c|c|c|c|}
\hline
{\bf type} & {\bf \stackanchor{patch size/}{stride}} & {\bf \stackanchor{output}{size}} &
{\bf depth} & {\bf $\#1{\times}1$} & {\bf \stackanchor{$\#3{\times}3$}{reduce}} & $\#3{\times}3$ &
{\bf \stackanchor{double $\#3{\times}3$}{reduce}} & {\bf \stackanchor{double}{ $\#3{\times}3$}} & {\bf Pool +proj} \\
\hline\hline
convolution* & $7{\times}7/2$ & $112{\times}112{\times}64$ & 1 & & & & & & \\
\hline
max pool & $3{\times}3/2$ & $56{\times}56{\times}64$ & 0 & & & & & & \\
\hline
convolution & $3{\times}3/1$ & $56{\times}56{\times}192$ & 1 & & 64 & 192 & & &  \\
\hline
max pool & $3{\times}3/2$ & $28{\times}28{\times}192$ & 0 & & & & & & \\
\hline
inception (3a) & & $28{\times}28{\times}256$ & 3 & 64 & 64 & 64 & 64 & 96 & avg + 32  \\
\hline
inception (3b) & & $28{\times}28{\times}320$ & 3 & 64 & 64 & 96 & 64 & 96 & avg + 64 \\
\hline
inception (3c) & stride 2 & $28{\times}28{\times}576$ & 3 & 0 & 128 & 160 & 64 & 96 & max + pass through \\
\hline
inception (4a) & & $14{\times}14{\times}576$ & 3 & 224 & 64 & 96 & 96 & 128 & avg + 128 \\
\hline
inception (4b) & & $14{\times}14{\times}576$ & 3 & 192 & 96 & 128 & 96 & 128 & avg + 128 \\
\hline
inception (4c) & & $14{\times}14{\times}576$ & 3 & 160 & 128 & 160 & 128 & 160 & avg + 128 \\
\hline
inception (4d) & & $14{\times}14{\times}576$ & 3 & 96 & 128 & 192 & 160 & 192 & avg + 128 \\
\hline
inception (4e) & stride 2 & $14{\times}14{\times}1024$ & 3 & 0 & 128 & 192 & 192 & 256 & max + pass through \\
\hline
inception (5a) & & $7{\times}7{\times}1024$ & 3 & 352 & 192 & 320 & 160 & 224 & avg + 128 \\
\hline
inception (5b) & & $7{\times}7{\times}1024$ & 3 & 352 & 192 & 320 & 192 & 224 & max + 128 \\
\hline
avg pool & $7{\times}7/1$ & $1{\times}1{\times}1024$ & 0 & & & & & & \\
\hline
  \end{tabular}
\end{center}
}
\caption{Inception architecture}
\label{fig-arch}
\end{figure*}

Figure~\ref{fig-arch} documents the changes that were performed compared to the architecture with respect to the GoogleNet archictecture. For the interpretation of this table, please consult \cite{inception}. The notable architecture changes compared to the GoogLeNet model include:
\begin{itemize}
\item The 5${\times}$5 convolutional layers are replaced by two consecutive 3${\times}$3 convolutional layers. This increases the maximum depth of the network by 9 weight layers. Also it increases the number of parameters by 25\% and the computational cost is increased by about 30\%.
\item The number 28${\times}$28 inception modules is increased from 2 to 3.
\item Inside the modules, sometimes average, sometimes maximum-pooling is employed. This is indicated in the entries corresponding to the pooling layers of the table.
\item There are no across the board pooling layers between any two Inception modules, but stride-2 convolution/pooling layers are employed before the filter concatenation in the modules 3c, 4e.
\end{itemize}
Our model employed separable convolution with depth multiplier $8$ on the first convolutional layer. This reduces the computational cost while increasing the memory consumption at training time.

\end{document}